\documentclass[letterpaper]{article} % DO NOT CHANGE THIS
\usepackage{aaai24}  % DO NOT CHANGE THIS
\usepackage{times}  % DO NOT CHANGE THIS
\usepackage{helvet}  % DO NOT CHANGE THIS
\usepackage{courier}  % DO NOT CHANGE THIS
\usepackage[hyphens]{url}  % DO NOT CHANGE THIS
\usepackage{graphicx} % DO NOT CHANGE THIS
\usepackage{natbib}  % DO NOT CHANGE THIS AND DO NOT ADD ANY OPTIONS TO IT
\usepackage{caption} % DO NOT CHANGE THIS AND DO NOT ADD ANY OPTIONS TO IT
\usepackage{algorithm}
\usepackage{algorithmic}

\usepackage{algorithm}
\usepackage{algorithmic}
\usepackage{booktabs}
\usepackage{amsmath}
\usepackage{subfigure}
\usepackage{amsfonts,bm}
\usepackage{multirow}
\usepackage{color}
\usepackage{braket}
\usepackage{mathtools}
\usepackage{enumerate}
\usepackage{enumitem}
\usepackage{empheq}
\usepackage{calc}
\usepackage{dsfont}
\usepackage{url}

\usepackage{float}

\usepackage[flushleft]{threeparttable}

\let\vec\textbf

\newcommand{\vx}{\vec{X}}
\newcommand{\vt}{\vec{T}}
\newcommand{\vy}{\vec{y}}
\newcommand{\vl}{\vec{L}}

\newcommand{\mf}{\mathcal{F}}
\newcommand{\ml}{\mathcal{L}}
\newcommand{\ms}{\mathcal{S}}
\newcommand{\mt}{\mathcal{T}}
\newcommand{\vw}{\vec{W}}
\newcommand{\vm}{\vec{M}}

\newcommand{\vtheta}{\mathbf{\Theta}}

\newcommand{\vij}{_i^{(j)}}

\newtheorem{definition}{Definition}
\usepackage{newfloat}
\usepackage{listings}
\DeclareCaptionStyle{ruled}{labelfont=normalfont,labelsep=colon,strut=off} % DO NOT CHANGE THIS
\floatstyle{ruled}
\newfloat{listing}{tb}{lst}{}
\floatname{listing}{Listing}
\title{Referee-Meta-Learning for Fast Adaptation of Locational Fairness}
\author {
    Weiye Chen\textsuperscript{\rm 1},
    Yiqun Xie\textsuperscript{\rm 1}\thanks{Corresponding author.},
    Xiaowei Jia\textsuperscript{\rm 2},
    Erhu He\textsuperscript{\rm 2},
    Han Bao\textsuperscript{\rm 3},
    Bang An\textsuperscript{\rm 3},
    Xun Zhou\textsuperscript{\rm 3}
}
\affiliations {
    \textsuperscript{\rm 1}University of Maryland\\
    \textsuperscript{\rm 2}University of Pittsburgh\\
    \textsuperscript{\rm 3}University of Iowa\\
    \{weiyec, xie\}@umd.edu, \{xiaowei, erh108\}@pitt.edu, \{han-bao, bang-an, xun-zhou\}@uiowa.edu
}

\usepackage{bibentry}
\usepackage{fancyhdr}
\usepackage{float}
\begin{document}

\maketitle
\thispagestyle{fancy}

\pdfoutput =1

\begin{abstract}
When dealing with data from distinct locations, machine learning algorithms tend to demonstrate an implicit preference of some locations over the others, which constitutes biases that sabotage the spatial fairness of the algorithm. This unfairness can easily introduce biases in subsequent decision-making given broad adoptions of learning-based solutions in practice. However, locational biases in AI are largely understudied. To mitigate biases over locations, we propose a locational meta-referee (Meta-Ref) to oversee the few-shot meta-training and meta-testing of a deep neural network. Meta-Ref dynamically adjusts the learning rates for training samples of given locations to advocate a fair performance across locations, through an explicit consideration of locational biases and the characteristics of input data. 
We present a three-phase training framework to learn both a meta-learning-based predictor and an integrated Meta-Ref that governs the fairness of the model.
Once trained with a distribution of spatial tasks, Meta-Ref is applied to samples from new spatial tasks (i.e., regions outside the training area) to promote fairness during the fine-tune step.
We carried out experiments with two case studies on crop monitoring and transportation safety, which show Meta-Ref can improve locational fairness while keeping the overall prediction quality at a similar level.
\end{abstract}

\section{Introduction}
\label{sec:intro}
Locational bias has been widely studied in many social sectors and linked with social disparities of various types \cite{fan2020spatial,kontokosta2021bias}, such as the impact of climate change and related disasters (e.g., floods, food shortage), resource distribution (e.g., subsidies in agriculture), and infrastructure quality and safety. 
With the increasing of adoptions of machine learning methods in broad domains (e.g., 
climate resilience, food security, public resource management), fairness issues associated with these prediction models have become a major subject, directly impacting the trust from the public and the sustained use of the systems in the long-term.
While fairness issues related to races or genders have been extensively examined in machine learning \cite{hardt2016equality, dwork2012fairness, zafar2017parity, pmlr-v80-agarwal18a, creager2019flexibly, kusner2017counterfactual}, very few studies have attempted to consider locational fairness.
The lack of consideration for locational fairness in machine learning applications may result in unintended consequences (e.g., biased resource distribution). In this study, we exemplify this issue by examining two important social problems:
(1) Agricultural monitoring: The climate change and the growing population have raised alarms and attention on global food security (e.g., G20's GEOGLAM initiative). 
With the broad deployment of machine learning in satellite-based crop monitoring (e.g., NASA Harvest), it is critical to explicitly consider locational fairness in mapping results, informing key decisions such as subsidy distribution or farm insurance \cite{bailey2010remote,national2018improving}.
(2) Transportation safety:
Given the complexity of traffic accident risk estimation, machine learning algorithms have been increasingly used to account for heterogeneous information from diverse sources. However, without the awareness of locational fairness, these methods can produce biased risk maps, further adding bias in investments for infrastructure improvements \cite{kontokosta2021bias, Bednarek_Boyce_Sileo_2022}.

We aim to create a new meta-learning framework that explicitly models locational fairness and enables rapid adaptation of fairness to new locations within different regions (e.g., different cities).
Unlike traditional fairness formulation using pre-defined groups such as races and genders, locational fairness faces more challenges when being transferred between training and test data.
First, in traditional fairness definitions, the groups considered in fairness evaluation are the same in the training and test datasets. In contrast, locational fairness often deals with entirely different sets of locations in the training region (e.g., city A) and test region (e.g., city B). Second, the change of spatial regions between training and test also introduces distribution shifts in the data. 
Third, fields such as agricultural monitoring require labor-intensive field survey, resulting in scarce availability of labeled data. Addressing these challenges requires the meta-learning model to learn the initial weights that not only can quickly adapt to the new distribution, but also adapt to an unknown fairness criterion (i.e., fairness defined on a new set of locations), from limited amount of labeled data.
Several directions have been explored in related work:

\noindent\textbf{Fair learning:} Fairness-aware learning formulations have been extensively studied and most existing works focus on pre-defined groups \cite{mehrabi2021survey}. A mainstream direction is to minimize the correlation between learned features with sensitive attributes, such as gender or race. The approaches include sensitive information encryption or removal \cite{kilbertus2018blind,johndrow2019algorithm}, feature decorrelation \cite{zhao2022towards}, agnostic representation learning \cite{creager2019flexibly, morales2020sensitivenets}, representation neutralization \cite{du2021fairness}, regularization \cite{yan2019fairst}, and so on.
However, these methods do not consider the scenario faced in this problem, where the groups represented by locations involved in fairness evaluation are different from training to test.

\noindent\textbf{Locational fairness:} Recent studies \cite{xie2022fairness,he2022sailing,he2023physics} examined fairness formulations with respect to locations, and they focus on the case where space partitions are used for fairness evaluation. Similarly, they only consider problems where the spatial region remains the same from training to testing, and cannot address the issue of non-stationary groups.

\noindent\textbf{Domain shifts:}
Domain adaptation methods mitigate covariance shift and learn invariant domains to reduce the effects of distribution shifts on model bias \cite{singh2021fairness}. Sample-reweighting and self-training approaches \cite{bickel2007discriminative,an2022transferring, he2023physics} also aim to reduce the distribution gap between training and testing sets by assigning higher weights to samples more similar to test samples feature-wise, or include high-confidence pseudo-labels on test samples during training. 
In addition, heterogeneity-aware learning tackles variability by data partitioning and network branching \cite{xie2021statistically,xie2023harnessing}.
While these methods address distribution shifts, they also do not consider the changes of groups (locations) between training and test for fairness applications.

\noindent\textbf{Meta-learning:}
Model-agnostic meta learning (MAML)’s gradient-by-gradient training allows it to learn an initial model that can be quickly fine-tuned to the test data with only a small number of observations \cite{finn2017model,ren2018learning,xie2023harnessing, chen2023physics}. Recent developments have also started exploring the use of MAML in fairness-aware learning \cite{zhao2020primal,zhao2022adaptive}.
These methods enable a fair model’s prediction to remain independent from the sensitive attributes, and can let it adapt to changing distributions. However, they similarly have not considered the case where training and test sets have completely different groups (locations). Moreover, we focus on a different class of fairness definitions -- prediction quality parity instead of protected attribute de-correlation (though both are commonly used standard definitions) \cite{zhang2018mitigating, du2020fairness} -- for our targeted applications.

We propose a referee-meta-learning framework to address the challenges. Our contributions are:
\begin{itemize}[noitemsep,nolistsep]
    \item We propose a locational meta-referee (Meta-Ref) which learns to dynamically adjust learning rates of data samples in a task to make the prediction model fairer for samples at different locations after the gradient updates.
    \item We propose a three-phase training framework to update parameters of Meta-Ref and its corresponding prediction model using a distribution of spatial tasks.
    \item We experiment with real-world data for satellite-based crop classification and traffic accident risk estimation. 
\end{itemize}
Our results on crop monitoring and transportation safety show that Meta-Ref can effectively improve fairness over locations in new test regions while keeping aggregated global performances similar to the baselines.

\section{Concepts and Problem Formulation}
The goal of this work is to mitigate locational biases in the prediction results, i.e., to reduce the variation of model performances, or \textbf{prediction quality disparity} \cite{du2020fairness}, over locations in a spatial region.

\begin{definition}
    \label{def:loc}
    \textup{
        A \textbf{location} $i$ is defined as a specific point or region in the geographical space, with $s_i$ representing all data points associated with the location $i$. A data point can be an one-dimensional vector, a time-series, an image, etc.
    } 
\end{definition}

\begin{figure*}[h]
	\centering\includegraphics[width=0.98\textwidth]{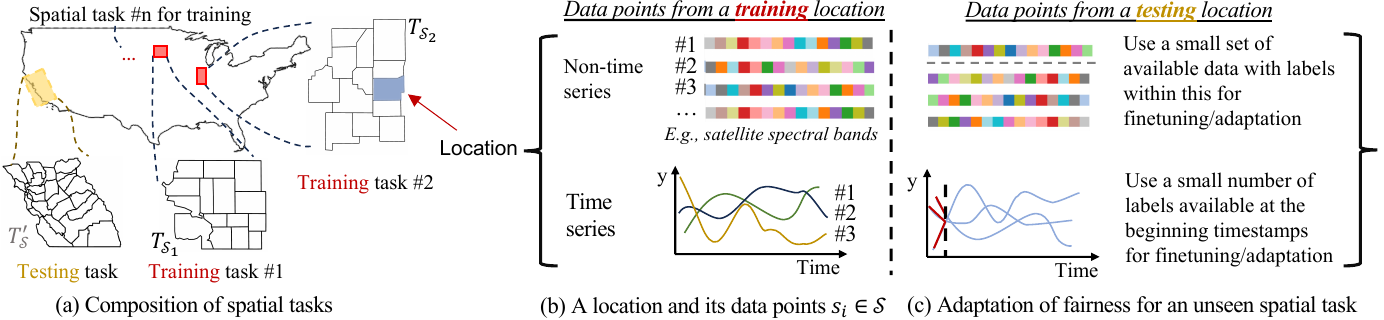}
	\caption{
	   Illustrative examples of spatial tasks for training and testing, their encompassing locations and data points, as well as the data requirements for finetuning/adaption for spatial tasks from test locations. 
	}
	\label{fig:location}
\end{figure*}

\begin{definition}\label{def:fair}
\textup{
    \textbf{Locational fairness}, $\ml_{fair}$, measures the locational biases of a deep neural network $\mf$ using the performance scores $\mathcal{M} = \{ m_{s_i}\}$ for data points from a set of distinct locations $\ms = \{s_i\}$: 
    \begin{equation}\label{eq:fair}
        \begin{split}
        \ml_{fair}(\ms) = \sqrt{\frac{1}{|\ms|} \sum_{s_i \in \ms} ( m_{s_i} - \hat{\mathcal{M}} )^2}
        \end{split}
    \end{equation}
    where $\hat{\mathcal{M}}$ is the average performance for all data points.
}
\end{definition}

\begin{definition}\label{def:task}
\textup{
A \textbf{spatial task} $T_\ms$ refers to the set of geo-located data points $\ms$ in a study area of interest, where a deep neural network $\mf$ with parameters $\vtheta$ is learned to make predictions.
$T_\ms$ also defines the set of locations where locational fairness is evaluated.}
\end{definition}

Fig. \ref{fig:location}(a) shows an example of sampling spatial tasks for training and testing, with counties as distinct locations. 
Fig. \ref{fig:location}(b) illustrates that a location may be associated with non-time series and/or time series data.  A standard machine learning algorithm does not consider locational fairness, potentially resulting in an unfair distribution of prediction quality scores across the spatial task. Conversely, we expect a fairness-driven algorithm to enhance parity among locations in terms of prediction quality scores.

\noindent\textbf{Problem Formulation.}
Given a set of spatial tasks $\{T_{\ms_1}, T_{\ms_2}, ...\}$ with associated features $\vx$ and labels $\vy$ from training locations, we aim to train a deep neural network $\mf_\vtheta(\cdot)$ with awareness of locational fairness (Eq. (\ref{def:fair})). The goal is that $\mf_\vtheta(\cdot)$ can be quickly adapted to a new spatial task $T_{\ms'}$ from test locations, where $T_{\ms'}\cap \{T_{\ms_1}, T_{\ms_2}, ...\} = \phi$, using only a small amount of test samples $\vx'$ and $\vy'$ (Fig. \ref{fig:location}(c)). The adaptation should consider both the prediction and fairness objectives.

\section{Method} 

In this section, we introduce our locational meta-referee (Meta-Ref). Meta-Ref works in conjunction with any neural network-based prediction models and dynamically assigns learning rates to mini-batches to enforce fairness. Trained using a meta-learning framework, Meta-Ref can adapt to various spatial tasks and can be fine-tuned for unseen ones. We will provide details on its structure as well as its training and transfer strategies in the following sections.

\subsection{Model-Agnostic Meta-Learning}

Model-Agnostic Meta-Learning (MAML) \cite{finn2017model} is a scheme which trains a generalizable model that can be quickly adapted to new tasks through gradient-by-gradient update strategies. It moderates and learns through gradient updates of over a set of tasks and finds a gradient leading to better generalization. The goal of MAML is to learn an initial model from a distribution of tasks such that the initial model can be quickly fine-tuned to the optimal parameters of individual tasks using only a few samples. Given a distribution of tasks $ \{T_i | T_i \sim p(\mt) \}$, each gradient update in MAML is given by:
\begin{equation}\label{eq:MAML_in}
    \vtheta_i^{\prime} \leftarrow \vtheta - \beta \nabla_{\vtheta} \ml(\vx^{(i)}, \mf_{\vtheta}, \vy^{(i)})
\end{equation}
\begin{equation}\label{eq:MAML_out}
    \vtheta \leftarrow \vtheta - \alpha \sum_{T_i \sim p(\mt)} \nabla_{\vtheta} \ml(\Tilde{\vx}^{(i)}, \mf_{\vtheta_i^{\prime}}, \Tilde{\vy}^{(i)})
\end{equation} 
where $\vtheta_i^{\prime}$ represents temporary parameters of the deep neural network $\mf$ for a task $T_i$; $\alpha$ and $\beta$ are the hyper-parameters of the step size of gradient updates; $\vx^{(i)}$ and $\vy^{(i)}$ are the training mini-batches, and $\Tilde{\mathbf{X}}^{(i)}$ and $\Tilde{\mathbf{y}}^{(i)}$ are the validation mini-batches for task $T_i$, respectively. 

This combination of task-specific and global gradient update simulates the scenarios encountered in testing, where we are given one initial model and aim to reach good performance of a new task after updating with a mini-batch. Thus, using the sequence of gradient updates (gradients of gradients), MAML learns a set of parameters that are not necessarily optimal for any given task, but can be quickly adapted to one specific task with a small number of points.
In this work, we define the tasks using spatial tasks $\{T_\ms\}$, which contain geo-located data points from different spatial regions.

\begin{figure*}[h]
	\centering
      \includegraphics[width=0.85\textwidth]{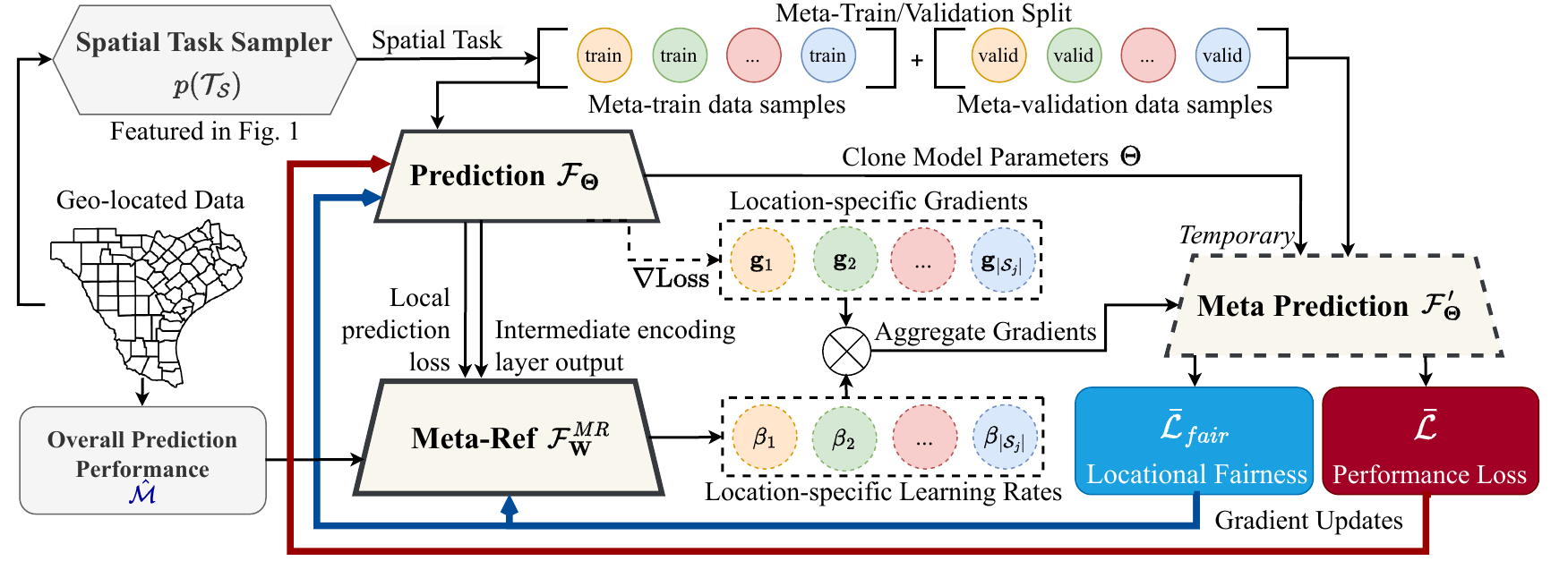}
	\caption{
	An illustration of the training framework to enforce locational fairness with Meta-Ref. }
	\label{fig:meta-ref}
\end{figure*}

\subsection{A Locational Meta-Referee}
We propose a locational meta-referee (Meta-Ref), which aims to adapt fairness learned from a distribution of spatial tasks $\{T_{\ms_1}, T_{\ms_2}, ...\}$ in the training area to a new spatial task $T_{\ms'}$ in the test region. Fig. \ref{fig:meta-ref} provides an illustration of the framework during training. The main ideas of Meta-Ref are: 
\begin{itemize}[leftmargin=*]
    \item It learns to produce learning rates for different locations within each spatial task as a function of the locations' features and the prediction model's relative performances on them, using only a few observations. 
    \item The prediction model is optimized with variable learning rates dynamically estimated by Meta-Ref, with the goal of collaboratively minimizing the fairness loss over the locations in a given spatial task.
    \item To quickly adapt to a new spatial task with a different set of locations, Meta-Ref creates a distribution of spatial tasks during training by randomly generating local subsets of data points at different locations, where the training spatial tasks may overlap with each other.
\end{itemize}
In other words, Meta-Ref $\mf^{MR}$ with parameters $\vw$ is trained in coordination with the prediction model $\mf$ to learn a mechanism to enforce fairness on a given spatial task and make sure the mechanism is easily transferable across a distribution of spatial tasks.

Specifically, Meta-Ref takes three types of inputs: 
(1) Performance metrics (e.g., RMSE) of data sample generated by the current prediction model $\mf$'s parameters $\vtheta$, which help evaluate the current performance on the data points and their potential impact on the locational fairness;
(2) The global performance metrics to benchmark the level of locational fairness and convert absolute performance metrics to relative scores;
and (3) The encoding generated by $\mf$ over data samples. The encoding reflects the characteristics of samples that can better guide the learning rate estimation. For example, a large loss may not always entail a high learning rate as a sample may be a very difficult case, whose loss can hardly be further reduced without causing significant negative impacts on other samples.
Denoting the encoding process of the prediction model as $\mf^{enc}$, we consider the prediction model as $\mf(\cdot) = \mf^{dec}(\mf^{enc}(\cdot))$, where $\mf^{dec}(\cdot)$ is the decoder.
Meta-Ref explicitly considers the performance metric $m_i = M(\vx_i, \mf_{\vtheta}, \vy_i)$ yielded by the prediction model on data samples. The performance metrics are further standardized by subtracting the global performance $\hat{\mathcal{M}}$ to obtain relative performances, so that Meta-Ref becomes invariant of the state of the overall performance, making it more transferable. 
Formally, we represent Meta-Ref as a neural network $\mf^{MR}$ with parameters $\vw$, which outputs a fairness factor $\eta_i$ for each location $s_i$ in a spatial task $T_\ms$ using:
\begin{equation}\label{eq:meta-ref}
    \eta_i = \mf^{MR}_{\vw}(\mf^{enc}(\vx_i), M(\vx_i, \mf_{\vtheta}, \vy_i)) - \hat{\mathcal{M}})
\end{equation}
The fairness factors will be translated into learning rates during meta-training, detailed in the next section. 

\subsection{Three-Phase Training of Meta-Ref}

We apply MAML's gradient-by-gradient update strategies to train both Meta-Ref and the prediction model.
Below, we present a three-phase framework for the training process.

\subsubsection{Generation of spatial task distribution $\mt(\ms)$.}
Before the start of the three-phase training, it is important to first generate a diverse distribution of spatial tasks so that we can learn a more transferable initial set of parameters for the prediction model $\mf$ in conjunction with Meta-Ref.
We combine two strategies to generate spatial tasks for the distribution $\mt(\ms)$:
(1) Locations are grouped according to administrative boundaries (e.g., cities, states), zones, or their attributes (e.g., precipitation, altitude) to create spatial tasks. To enlarge the variety of spatial tasks, the elements of two spatial tasks do not need to be disjoint, so there's a possibility that $\exists \ms_i, \ms_j \in \{\ms\}: \ms_i \cap \ms_j \neq \emptyset$. 
(2) To create spatial tasks that are more distinctive from regular ones from the training area, we further include spatial tasks whose locations are randomly sampled from the entire training area.

\subsubsection{Phase 1: Prediction performance estimation.}

Prior to each epoch of training, we begin by sampling an array of spatial tasks $\vt = \{T_{\ms_i} \sim p(\mt(\ms))\}$.
The first step is to generate the necessary inputs for Meta-Ref according to Eq. (\ref{eq:meta-ref}). In this phase, Meta-Ref gains knowledge of how the prediction model performs on spatial tasks and assesses the level of locational biases in them. On a training mini-batch $(\vx\vij, \vy\vij)$ at location $s\vij$ in the spatial task $T_{\ms_j}$, we assess the prediction loss as $l\vij = \ml(\vx\vij, \mf_{\vtheta}, \vy\vij)$ and the corresponding metrics $m\vij = M(\vx\vij, \mf_{\vtheta}, \vy\vij)$.

We apply the same procedure on all locations in a spatial task $T_{\ms_j}$, getting an losses and metrics, $\vl^{(j)}$ and $\vm^{(j)}$:
\begin{equation}\label{eq:array_loss_met}
\begin{split}
    \kern -0.58em 
    \vl^{(j)} = \left[l\vij \bigg| i \in \left[1, |\ms_j|\right]\right],  \vm^{(j)} = \left[m\vij \bigg| i \in \left[1, |\ms_j|\right]\right]
\end{split}
\end{equation}
With loss values calculated for all locations, we compute gradients of each location as 
        $\mathbf{g}\vij = \nabla_{\vtheta} l\vij$.
Additionally, we evaluate the overall performance for all training data points regardless of their location origin $(\vx, \vy)$:
\begin{equation}\label{eq:overall_per_met}
    \hat{\mathcal{M}} = M(\vx, \mf_{\vtheta}, \vy)
\end{equation}

\subsubsection{Phase 2: Fairness-aware learning rate estimation.}

Using outputs $\vl^{(j)}$, $\vm^{(j)}$, $\mathbf{g}\vij$ and $\hat{\mathcal{M}}$ from Phase 1, Meta-Ref dynamically adjusts the step sizes of gradients associated with data points at different locations in a spatial task: 
\begin{equation}\label{eq:MAML_ref_in}
    \vtheta^{\prime(j)} \leftarrow \vtheta - \sum_{s_i \in \ms_j}\beta\vij\mathbf{g}\vij
\end{equation}
where $\beta\vij$ represents the learning rate of data points in location $s\vij$, assigned by Meta-Ref, and $\vtheta^{\prime(j)}$ is the temporary parameters of $\mf$ through updates on data points of all locations. 
Different from MAML (Eq. (\ref{eq:MAML_in})), step sizes are no longer identical over all data points in a mini-batch; instead, they become dependent on the locations and spatial tasks. 
Step sizes are assigned via translating Meta-Ref-generated fairness factors $    \mathbf{N}^{(j)} = \left[ \eta\vij \bigg| i \in \left[1, |\ms_j|\right]\right]$ with Eq. (\ref{eq:meta-ref}). Specifically, we perform the translation by standardizing the fairness factors from all locations within each spatial task into a stationary range to improve the stability of training:
\begin{equation}\label{eq:}
    \Tilde{\eta}\vij = \frac{\eta\vij - \min(\mathbf{N}^{(j)})}{\max(\mathbf{N}^{(j)}) - \min(\mathbf{N}^{(j)})}
\end{equation}
\begin{equation}\label{eq:lr_ff}
    \beta\vij = \Tilde{\eta}\vij \times (\beta^+-\beta^-) + \beta^-
\end{equation}
\noindent where $\eta\vij = \mf^{MR}_{\vw}\left(\mf^{enc}_{\vtheta}(\vx\vij), m\vij - \hat{\mathcal{M}}\right)$; and $\beta^+$ and $\beta^-$ represent the upper and lower limits of step size of the gradient update, respectively.

To stabilize training at the early stage, we constrain the variance of learning rates across data samples from different locations, $\text{var}\left(\left\{\beta\vij \big| i \in \left[1, |\ms_j|\right]\right\}\right)$, by adjusting $\beta^+$ and $\beta^-$. Given a baseline learning rate $\beta^0$ and a scaling factor $\rho$, we set the upper and lower bounds of $\beta\vij$ at iteration $t$ as $\beta^+ = \frac{1}{1 + \text{e}^{-t/\rho}} \cdot \beta^0$ and $\beta^- = \frac{\text{e}^{-t/\rho}}{1 + \text{e}^{-t/\rho}} \cdot \beta^0$. The gap between these bounds expands gradually through a sigmoid-shaped curve as training progresses, so after the early training stage learning rates approach a constant range, allowing more flexibility to improve fairness.

Finally, the location-dependent learning rates are applied via Eq. (\ref{eq:MAML_ref_in}) for gradient updates on the temporary prediction model parameters within the inner meta-training loop.

\subsubsection{Phase 3: Dual meta-updates.}
In this dual meta-update phase, 
we consider both the prediction performance and the locational fairness to make final gradient updates.
Specifically, we use two different losses on the validation data for a spatial task $T_{\ms_j}$, prediction loss $\bar{\ml}^{(j)}$ and locational fairness loss $\bar{\ml}^{(j)}_{fair}$, to meta-update the parameters of the prediction model and Meta-Ref. 
The prediction loss measures the collective performance of temporarily updated prediction model parameters $\Tilde{\eta}\vij$, and the locational fairness loss reflects the effectiveness on the coordination of Meta-Ref and the prediction model. 
We compute two losses using:
\begin{equation}\label{eq:pred_loss_sum}
    \begin{split}
        \bar{\ml}^{(j)} = \ml(\Tilde{\vx}^{(j)}, \mf_{\vtheta^{\prime(j)}}, \Tilde{\vy}^{(j)})
    \end{split}
\end{equation}
\begin{equation}\label{eq:fairnes_loss}
    \begin{split}
        \bar{\ml}^{(j)}_{fair} & = \ml_{fair}(\ms_j) = \sqrt{\frac{1}{|\ms_j|} \sum_{s_i \in \ms_j} (\Tilde{m}\vij - \hat{\mathcal{M}} )^2}
    \end{split}
\end{equation}
\noindent where $\Tilde{m}\vij = M(\Tilde{\vx}\vij, \mf_{\vtheta^{\prime(j)}}, \Tilde{\vy}\vij)$; ($\Tilde{\vx}^{(j)}$, $\Tilde{\vy}^{(j)}$) are validation data from the whole $\ms_j$; and ($\Tilde{\vx}\vij$, $\Tilde{\vy}\vij$) are validation mini-batch sampled from location $s_i$.

For the prediction loss $\bar{\ml}^{(j)}$, we use its gradients to update only the prediction model. For $\bar{\ml}^{(j)}_{fair}$, we use its gradients to update both the prediction model and Meta-Ref.
In this way, Meta-Ref focuses on the fairness side, and it coordinates with the prediction model to address both prediction performance and fairness:
\begin{equation}\label{eq:perf_pred}
        \vtheta \leftarrow \vtheta - \alpha_1 \nabla_{\vtheta}\bar{\ml}^{(j)}
\end{equation}
\begin{equation}\label{eq:fair_mr}
    \begin{split}
        \vw \leftarrow \vw - \alpha_2 \nabla_{\vw}\bar{\ml}^{(j)}_{fair}
    \end{split}
\end{equation} 
\begin{equation}\label{eq:fair_pred}
    \begin{split}
        \vtheta \leftarrow \vtheta - \alpha_3 \nabla_{\vtheta}\bar{\ml}^{(j)}_{fair}
    \end{split}
\end{equation}
where $\alpha_1$, $\alpha_2$, and $\alpha_3$ are learning rates set for the three meta-update operations, respectively. 
Through Eqs. (\ref{eq:pred_loss_sum} - \ref{eq:fair_pred}), we can see that each of the three-way meta-update involves gradients over gradients (since $\vtheta^{\prime(j)}$ remains in expanded forms of Eqs. (\ref{eq:perf_pred} - \ref{eq:fair_pred})). This makes the gradient updates consider the contribution of each location in a spatial task, and mimic the actual fine-tuning process during testing.
Algorithm \ref{alg:algorithm} summarizes the procedure of the three-phase training of Meta-Ref. 

\begin{algorithm}[h]
\caption{Three-Phase Training of Meta-Ref}
\label{alg:algorithm}
\textbf{Require}: $p(\mt_{\ms})$: distribution of spatial tasks \\ 
\textbf{Parameters}: $\alpha_1$, $\alpha_2$, $\alpha_3$, $\beta^0$, $\rho$
\begin{algorithmic}[1] %[1] enables line numbers
\small
\STATE sample batch of spatial tasks $\vt = \{T_{\ms_j} \sim p(\mt)\}$
\FORALL{$T_{\ms_j} \in \vt$}
    \STATE \textbf{[Phase 1]}
    \FORALL{$s\vij \in \ms_j$}
        \STATE Evaluate local prediction loss $l\vij = \ml(\vx\vij, \mf_{\vtheta}, \vy\vij)$
    \ENDFOR
    \STATE Evaluate global prediction loss $\hat{\mathcal{M}} = M(\vx, \mf_{\vtheta}, \vy)$
    \STATE \textbf{[Phase 2]}
    \STATE Assign step size $\beta\vij$ with Eqs. (\ref{eq:} - \ref{eq:lr_ff}) for $s\vij \in \ms_j$
    \STATE Update $\vtheta^{\prime(j)} \leftarrow \vtheta - \sum_{s_i \in \ms_j}\beta\vij\mathbf{g}\vij$ [Eq. \ref{eq:MAML_ref_in}]
    \STATE  \textbf{[Phase 3]}
    \STATE Evaluate $\bar{\ml}^{(j)}$ with Eq. (\ref{eq:pred_loss_sum})
    \STATE Evaluate $\bar{\ml}^{(j)}_{fair}$ with Eq. (\ref{eq:fairnes_loss})
    \STATE Update $\vtheta \leftarrow \vtheta - \alpha_1 \nabla_{\vtheta}\bar{\ml}^{(j)}$
    \STATE Update $\vw \leftarrow \vw - \alpha_2 \nabla_{\vw}\bar{\ml}^{(j)}_{fair}$
    \STATE Update $\vtheta \leftarrow \vtheta - \alpha_3 \nabla_{\vtheta}\bar{\ml}^{(j)}_{fair}$
\ENDFOR
\end{algorithmic}
\end{algorithm}

\subsection{Fine-Tuning on Test Region}

Meta-trained parameters of the prediction model are expected to demonstrate good generalizability but not necessarily optimal to any tasks. Though, we do not fine-tune Meta-Ref, considering that it is not directly related to the prediction performance, while also avoiding overfitting. 

Given a spatial task in the test area $T_{\ms'}$, where $T_{\ms'} \cap \{T_{\ms_1}, T_{\ms_2}, ...\} = \phi$, we fine-tune the test data $\vx'$ and $\vy'$ in a slightly different fashion than three-phase meta-updates. The fine tuning has two phases: 1) Prediction performance estimation, and 2) Meta-Ref-guided optimization. 

Phase 1 of fine-tuning is similar to the Phase 1 of training Meta-Ref on a spatial task, with the differences being that the input data are from different regions, and that the overall performance metrics $\hat{\mathcal{M}}$ remains being calculated from the training data. Formally, we evaluate the local prediction loss on every location $s_i' \in \ms'$ with $l_i' = \ml(\vx_i',  \mf_{\vtheta}, \vy_i')$ and the global overall performance metrics $\hat{\mathcal{M}} = M(\vx, \mf_{\vtheta}, \vy)$. 

Phase 2 follows with the assignment of learning rates of all locations within this spatial task, following Eqs. (\ref{eq:} - \ref{eq:lr_ff}) using Meta-Ref, producing $\beta_i'$ for each location in the test region. Then we optimize $\vtheta$ with $\vtheta \leftarrow \vtheta - \sum_{s_i' \in \ms'} \beta_i'\nabla_{\vtheta}l_i'$. 
This two-phase fine-tuning effectively simulates the behavior of Eq. (\ref{eq:MAML_ref_in}) where we update the prediction model with Meta-Ref-assigned learning rates.

\section{Experiments}

\subsection{Case Study Datasets}

\textbf{Satellite-based crop classification:}
Crop mapping is important for various downstream tasks including acreage estimation, subsidy distribution and farm insurance.
Our study area is a $\sim$6700 km$^2$ region in Central Valley, California, which is a major region in the US for walnut plantation. 
The satellite imagery we use is from Sentinel-2 multispectral data. As the 10 spectral bands we use from Sentinel-2 have different spatial resolution (e.g., 10m, 20m), we sample all bands to 20m (with an image tile size of $4096\times4096$), a common choice in applications. The image tile was captured in August 2018. The labels are from the USDA Crop Data Layer (CDL)~\cite{USDA_cdl}. For walnut plantation mapping, we preprocess the labels into binary walnut and non-walnut classes.
Since the fairness paradigm is based on prediction quality parity, fairness calculation needs the performances from the locations as inputs.
For classification, this requires a certain level of aggregation (e.g., F1 or accuracy is not meaningful for an individual point). In our experiment, locations are thus represented by $128\times128$ non-overlapping local patches from the image tile instead of individual pixels.
We use a 50\% by 50\% train-test split for the locations.
In a test location, 5\% of randomly sampled data points are used for fine-tuning. Each spatial task $\ms_i$ is randomly sampled from training or test locations covered by a random $1280\times1280$ window (a $\sim$25km$\times$25km region), with the number of locations ranging from 10 to 15 per spatial task. 

\noindent\textbf{Traffic accident risk estimation:}
Location-based biases in transportation safety estimation can further lead to biases for investment distribution for infrastructure improvements.
We use the Iowa traffic accident record dataset shared by \cite{an2022hintnet},
which contains 3 years of traffic accident records and 47 related factors.
The dataset has a daily temporal resolution and was spatially aggregated into grid cells, and the total grid size is $64\times128$ \cite{an2022hintnet}. 
We use each cell as a location for this regression problem.
We partition the dataset into 8-week moving windows, where we use factors in the first 7 weeks to predict the average daily count of accidents in the eighth week.
Similarly, we use a 50\% by 50\% train-test split for the locations. In a test location, only the first 5\% of moving windows are used for fine-tuning.
Each spatial task is randomly sampled from training or testing locations in a randomly-selected $32\times32$ window, with 10 to 15 locations per spatial task.

\subsection{Methods for Comparison}

We evaluate the following methods in terms of prediction performance and, particularly, locational fairness: 
(1) \textbf{DNN} and \textbf{LSTM}: Plain baselines of neural network, including fully-connected deep neural network for crop classification with image snapshots (non-time-series) and Long-short-term memory (LSTM) model for traffic accident prediction with time series inputs. 
(2) \textbf{Reg}: DNN or LSTM with an additional regularization term (i.e., variance) to enforce locational fairness, a common strategy for prediction quality parity \cite{kamishima2011fairness,yan2019fairst}.
(3) \textbf{Adv}: An adversarial training method to learn a location-neutral representation of data samples, inspired by \cite{zhang2018mitigating, alasadi2019toward}.
(4) \textbf{Domain}: We add domain adaptation to plain baselines (\textbf{Dom-DNN}, \textbf{Dom-LSTM}) and \textbf{Reg} (\textbf{Dom-Reg}), where a discriminator is used to learn domain-invariant features. This can help bridge potential domain gaps between tasks in training and testing.
(5) \textbf{Bi-Lvl}: A recent state-of-the-art for improving locational fairness by adjusting fairness based on relative performances among samples \cite{xie2022fairness, he2023physics}. It is designed for tasks in the same region and does not consider generalization to new tasks.
(6) \textbf{MAML}: Model-agnostic meta-learning \cite{finn2017model}, designed for fast adaption to new tasks. MAML does not consider the fast adaptation of locational fairness. 
(7) We compare the aforementioned methods against \textbf{Meta-Ref}: Our proposed approach. More details can be found in the technical appendix.

\subsection{Evaluation Metrics}

To evaluate the prediction quality of a test spatial task $T_{\ms^{\prime}}$ from crop classification, we use F1-score $p_i$ to account for the prediction quality for data points at location $s_i \in \ms^{\prime}$. 
For a regression task in traffic accident risk estimation, we adopt the 
root mean squared error (RMSE) for $p_i$. 
We assess the locational fairness (LF) on the test spatial task from the evaluation metrics at each location by taking their standard deviation. 
Some methods producing poor prediction quality might achieve better locational fairness on some spatial tasks.
Therefore, we also include an adjusted locational fairness (ALF) metric to account for differences in prediction quality when evaluating fairness. 
Instead of using the mean prediction performance from the current method to calculate the standard deviation, we use the best prediction performance for this task among all methods as the reference mean, denoted as $p^*$.
Then the adjusted fairness score for a spatial task is defined as the average deviation of the prediction quality $\{p_i\}$ from this reference $p^*$:
\begin{equation}
ALF = \sqrt{\frac{1}{|\ms^{\prime}|} \sum_{s_i \in \ms^{\prime}} (p_i - p^*)^2} 
\end{equation}
By setting the mean to the best prediction performance among all methods, methods that trade performance for fairness (i.e., low variance among $\{p_i\}$ but higher distance to $p^*$) will be penalized and produce worse scores in ALF.

\begin{table*}[t]
  \centering
  \small
  \begin{tabular}{|c||c|cc||c|cc||c|cc|}
  \hline
  & \multicolumn{3}{c||}{{{Task set 1}}}  & \multicolumn{3}{c||}{{{Task set 2}}} &\multicolumn{3}{c|}{{{Task set 3}}} \\ \cline{2-10}
  & \phantom{---}{F1}\phantom{---}  & \phantom{--}{LF}\phantom{--}    & \phantom{--}{ALF}\phantom{--}   & \phantom{---}{F1}\phantom{---}  & \phantom{--}{LF}\phantom{--}    & \phantom{--}{ALF}\phantom{--}   & \phantom{---}{F1}\phantom{---}  & \phantom{--}{LF}\phantom{--}    & \phantom{--}{ALF}\phantom{--}     \\\hline
  DNN     & 0.681 & 0.130 & 0.138 & 0.680 & 0.134 & 0.141 & 0.677 & 0.133 & 0.139           \\
  Reg      & 0.680 & 0.129 & 0.137 & 0.680 & 0.134 & 0.141 & 0.676 & 0.133 & 0.139           \\
  Adv      & 0.660 & 0.130   & 0.147     & 0.640 & 0.127   & 0.158   & 0.639 & 0.139      & 0.158        \\
  Dom-DNN  & 0.660 & 0.133 & 0.148 & 0.652 & 0.148 & 0.164 & 0.654 & 0.143 & 0.157         \\
  Dom-Reg  & 0.664 & 0.130 & 0.144 & 0.656 & 0.143 & 0.159 & 0.659 & 0.138 & 0.151         \\
  Bi-Lvl   & 0.657 & 0.128 & 0.147 & 0.637 & 0.126 & 0.156 & 0.641 & 0.133 & 0.157         \\
  MAML     & 0.719 & 0.119 & 0.119 & 0.718 & 0.124 & 0.124 & 0.714 & 0.126 & 0.126         \\
  Meta-Ref & 0.716 & \textbf{0.107} & \textbf{0.107} & 0.710 & \textbf{0.114} & \textbf{0.115} & 0.706 & \textbf{0.118} & \textbf{0.119}  \\ \hline
  \end{tabular}
  \caption{Average metrics for satellite-based crop classification on 90 spatial tasks from test locations.}
  \label{tb:perf_cls}
  \end{table*}

\begin{table*}[t]
\centering
\small
\begin{tabular}{|c||c|cc||c|cc||c|cc|}
  \hline
  
    & \multicolumn{3}{c||}{{{Task set 1}}}  & \multicolumn{3}{c||}{{{Task set 2}}} &\multicolumn{3}{c|}{{{Task set 3}}}\\ \cline{2-10}
  & \phantom{-}{RMSE}\phantom{-}  & \phantom{--}{LF}\phantom{--}    & \phantom{--}{ALF}\phantom{--}   & \phantom{-}{RMSE}\phantom{-}  & \phantom{--}{LF}\phantom{--}    & \phantom{--}{ALF}\phantom{--}   & \phantom{-}{RMSE}\phantom{-}  & \phantom{--}{LF}\phantom{--}    & \phantom{--}{ALF}\phantom{--}   \\\hline
  LSTM     & 0.139 & 0.115          & 0.116          & 0.137 & 0.102          & 0.104          & 0.150 & 0.121          & 0.125              \\
  Reg      & 0.146 & 0.126          & 0.128          & 0.139 & 0.107          & 0.110          & 0.138 & 0.108          & 0.109              \\
  Adv      & 0.141 & 0.129          & 0.133          & 0.140 & 0.111          & 0.114          & 0.149 & 0.110          & 0.114             \\
  Dom-LSTM & 0.137 & 0.110          & 0.114          & 0.132 & 0.089          & 0.090          & 0.135 & 0.088          & 0.089             \\
  Dom-Reg  & 0.139 & 0.110          & 0.115          & 0.137 & 0.089          & 0.092          & 0.136 & 0.088          & 0.090             \\
  Bi-Lvl   & 0.149 & 0.133          & 0.135          & 0.146 & 0.108          & 0.112          & 0.150 & 0.110          & 0.114             \\
  MAML     & 0.134 & 0.108          & 0.110          & 0.133 & 0.084          & 0.087          & 0.137 & 0.087          & 0.089             \\
  Meta-Ref & 0.131 & \textbf{0.102} & \textbf{0.103} & 0.136 & \textbf{0.081} & \textbf{0.083} & 0.138 & \textbf{0.085} & \textbf{0.086}     \\ \hline
  \end{tabular}
\caption{Average metrics for traffic accident risk estimation on 90 spatial tasks from test locations.}
\label{tb:perf_regr}
\end{table*}

\subsection{Results}

\begin{figure}[b]
    \centering
    \includegraphics[width=1.0\linewidth]{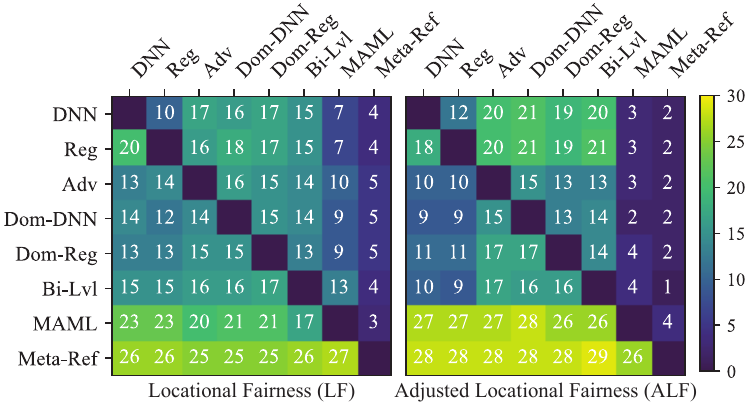}
    \caption{
       A pairwise comparison matrix for all methods (This is an example from task set 1 for crop classification with 30 tasks in total).
    }
    \label{fig:comp_mat_cls}
\end{figure}

\noindent\textbf{Training:}
We sample 1000 spatial tasks from training locations of each dataset. 
Then we sample another 90 spatial tasks from testing locations from each dataset and split them into three folds for testing. 
All methods are fine-tuned in each of the 90 spatial tasks in test regions for both datasets.

\begin{figure}[t]
    \centering
    \includegraphics[width=1\linewidth]{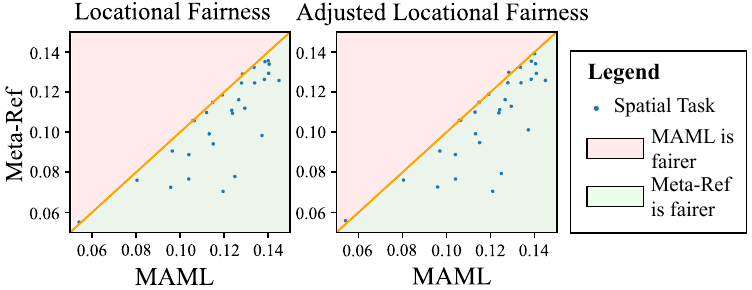}
    \caption{
        Comparison between MAML and Meta-Ref on fairness metrics among different spatial tasks (an example from task set 1 for crop classification with 30 tasks in total). 
    }
    \label{fig:maml_mr_scatter}
\end{figure}

\noindent\textbf{Comparison to baselines:}
Table \ref{tb:perf_cls} and \ref{tb:perf_regr} show the average performance of different models on two datasets. MAML and Meta-Ref have similar performance in terms of prediction metrics on both datasets. In crop classification tasks, these two methods outperform other baselines by significant margins in both performance and fairness measures. In traffic accident risk estimation dataset, Domain methods are close to MAML in both performance and fairness but never surpass MAML. Meta-Ref has reliably demonstrated a lead over MAML in both locational fairness (LF) and adjusted locational fairness (ALF) on both datasets. 
The leading positions of Meta-Ref in ALF across both datasets suggest that Meta-Ref does not demonstratively sacrifice its prediction performance for fairness. Meta-Ref tends to generate fairer results even when setting the benchmark to the best-performing model, if not Meta-Ref, for most test tasks. 
In Fig. \ref{fig:comp_mat_cls} and also in the technical appendix, we demonstrate pairwise comparison matrices, where each element indicates the number of spatial tasks where the row method has lower fairness metrics (LF or ALF) than column method. It shows that Meta-Ref maintain better fairness on most spatial tasks compared to baselines. 
In addition, as shown in Fig. \ref{fig:maml_mr_scatter}, Meta-Ref has demonstrated fairer predictions than MAML on most tasks, confirming its effectiveness.

\noindent\textbf{Sensitivity analysis:}
The training of Meta-Ref relies on three outer gradient updates (Eqs. (\ref{eq:perf_pred}, \ref{eq:fair_mr}, \ref{eq:fair_pred})) to coordinate the performance loss and fairness loss with their impacts on prediction and meta-referee. 
To demonstrate the effectiveness of three outer gradient updates in Meta-Ref, we further conduct an ablation study with the following models: (1) \textbf{MR-P2P}: Meta-Ref without applying performance gradient to the prediction model (P2P, Eq. (\ref{eq:perf_pred})); (2) \textbf{MR-F2M}: Meta-Ref without applying fairness gradient to the meta-referee (F2M, Eq. (\ref{eq:fair_mr})); and (3) \textbf{MR-F2P}: Meta-Ref without applying fairness gradient to the prediction model (F2P, Eq. (\ref{eq:fair_pred})). 
The experiment results suggest that all gradient updates are essential in the training of Meta-Ref. Without P2P, the model fails to generalize with spatial tasks. Without F2M, meta-referee is no longer updated during training, producing arbitrary locational fairness factors that disrupt the training of prediction model. Trimming F2P from Meta-Ref has the smallest impact among three ablation models, since while MR-F2P still underperforms full MAML and Meta-Ref, it has better prediction and location fairness metrics than other baselines. The technical appendix provides more details about this analysis as well as the robustness of hyper-parameters and results.

\section{Conclusions}

This study presented Meta-Ref, a locational fairness meta-referee to address the implicit biases of a neural network on geo-located data. Trained using a meta-learning framework, Meta-Ref attributes the scale of locational biases with characteristics of data and training performance dynamics, and assigns learning rates to data points from different locations. Meta-Ref can be applied to a fine-tuned prediction model on location sets that have never been seen during meta-training. 
Case studies on crop monitoring and transportation safety showed that Meta-Ref can effectively improve locational fairness.
Our future work will explore domain customizations to facilitate implementation in real practices.

\section*{Acknowledgements}
This material is based upon work supported by the National Science Foundation under Grant No. 2105133, 2126474 and 2147195; NASA under Grant No. 80NSSC22K1164 and 80NSSC21K0314; USGS grants G21AC10207, G21AC10564, and G22AC00266; Google's AI for Social Good Impact Scholars program; the DRI award and the Zaratan supercomputing cluster at the University of Maryland; Pitt Momentum Funds award and CRC at the University of Pittsburgh; and the ISSSF grant from the University of Iowa.

\bigskip

\bibliography{aaai24}

\clearpage
\appendix

\section{Appendix}
\setcounter{table}{0}
\renewcommand{\thetable}{A\arabic{table}}

\setcounter{figure}{0}
\renewcommand{\thefigure}{A\arabic{figure}}

\section{Additional Details about Experimental Setup}

In order to reproduce the experimental results that are demonstrated in the main document as well as those in addition shown in this technical appendix, there are several hyperparameters mentioned in the main document in which a reader might be interested. We sample 1000 spatial tasks from the training locations of each dataset. To stabilize training of Meta-Ref at the early stage, we gradually increase the gap between step size among distinct locations in the spatial task in Phase 2. The baseline learning rate for inner MAML update $\beta^0$ we used in phase 2 is $10^{-4}$, and we set the scaling factor $\rho$ for the widening gap to be $50$. The training of Meta-Ref relies on three outer gradient updates (Eqs. (\ref{eq:perf_pred}, \ref{eq:fair_mr}, \ref{eq:fair_pred})) to coordinate the performance loss and fairness loss with their impacts on prediction and meta-referee. Using the gradients of performance loss to update the prediction model (Eq. (\ref{eq:perf_pred})) is standard in the baseline MAML, while using gradients of fairness loss to update the meta-referee (Eq. (\ref{eq:fair_mr})) and the prediction model (Eq. (\ref{eq:fair_pred})) is unique in our proposed model. We set the learning rate for three respective updates as the following: $\alpha_1 = 10^{-3}, \alpha_2 = \alpha_3 = 10^{-4}$. More experiments regarding these parameters can be found in the sensitivity analysis result in this document. For crop classification dataset, we employed a 10-layer fully connected neural network with residual connections. As for the traffic accident estimation dataset, we attach additional fully-connected layers before LSTM layers for feature learning. Both DNN and LSTM were trained using different candidate methods described in the main documents. 

Our implementation of the proposed method uses TensorFlow 2.10 with CUDA. The experiments were primarily conducted on a Windows 10 workstation with an Intel Xeon W-2265 processor, 64GB of RAM, and one NVIDIA A4500 graphic processing unit. The code for this implementation is attached in the Code \& Data Appendix to this document.

\begin{table*}[h]
  \centering
  \small
  \begin{tabular}{|c||c|cc||c|cc||c|cc|}
  \hline
  & \multicolumn{3}{c||}{{{Task set 4}}}  & \multicolumn{3}{c||}{{{Task set 5}}} &\multicolumn{3}{c|}{{{Task set 6}}} \\ \cline{2-10}
  & \phantom{---}{F1}\phantom{---}  & \phantom{--}{LF}\phantom{--}    & \phantom{--}{ALF}\phantom{--}   & \phantom{---}{F1}\phantom{---}  & \phantom{--}{LF}\phantom{--}    & \phantom{--}{ALF}\phantom{--}   & \phantom{---}{F1}\phantom{---}  & \phantom{--}{LF}\phantom{--}    & \phantom{--}{ALF}\phantom{--}     \\\hline
DNN      &  0.705  & 0.127    & 0.129          & 0.700 & 0.125          & 0.127          & 0.703 & 0.125          & 0.130          \\
Reg      &  0.704  & 0.126    & 0.128          & 0.700 & 0.124          & 0.127          & 0.703 & 0.125          & 0.130          \\
Adv      &  0.690  & 0.129    & 0.135  & 0.679 & 0.118   & 0.127    & 0.693 & 0.125 & 0.135  \\
Dom-DNN  &  0.678 & 0.137          & 0.148          & 0.682 & 0.131          & 0.138          & 0.682 & 0.131          & 0.142          \\
Dom-Reg  &  0.680 & 0.136          & 0.146          & 0.682 & 0.130          & 0.136          & 0.685 & 0.129          & 0.139          \\
Bi-Lvl   &  0.678 & 0.121          & 0.132          & 0.676 & 0.118          & 0.128          & 0.687 & 0.120          & 0.131          \\
MAML     &  0.723 & 0.124          & 0.124          & 0.719 & 0.123          & 0.123          & 0.721 & 0.121          & 0.123          \\
Meta-Ref &  0.712 & \textbf{0.112} & \textbf{0.114} & 0.709 & \textbf{0.111} & \textbf{0.112} & 0.715 & \textbf{0.113} & \textbf{0.116} \\ \hline

  \end{tabular}
  \caption{Average metrics for satellite-based crop classification on another 90 spatial tasks from test locations.}
  \label{tb:perf_cls_2}
  \end{table*}

\begin{table*}[h]
\centering
\small
\begin{tabular}{|c||c|cc||c|cc||c|cc|}
  \hline
    & \multicolumn{3}{c||}{{{Task set 4}}}  & \multicolumn{3}{c||}{{{Task set 5}}} &\multicolumn{3}{c|}{{{Task set 6}}}\\ \cline{2-10}
  & \phantom{-}{RMSE}\phantom{-}  & \phantom{--}{LF}\phantom{--}    & \phantom{--}{ALF}\phantom{--}   & \phantom{-}{RMSE}\phantom{-}  & \phantom{--}{LF}\phantom{--}    & \phantom{--}{ALF}\phantom{--}   & \phantom{-}{RMSE}\phantom{-}  & \phantom{--}{LF}\phantom{--}    & \phantom{--}{ALF}\phantom{--}   \\\hline
LSTM     &  0.130 & 0.099          & 0.100          & 0.124 & 0.087          & 0.088          & 0.134 & 0.112          & 0.114          \\
Reg      &  0.136 & 0.104          & 0.106          & 0.128 & 0.088          & 0.090          & 0.133 & 0.117          & 0.119          \\
Adv      &  0.136 & 0.109           & 0.115         & 0.132 & 0.106          & 0.108          & 0.141 & 0.113 & 0.117\\
Dom-LSTM &  0.126 & 0.097          & 0.099          & 0.128 & 0.093          & 0.096          & 0.128 & 0.093          & 0.094          \\
Dom-Reg  &  0.130 & 0.092          & 0.095          & 0.133 & 0.095          & 0.097          & 0.132 & 0.088          & 0.090          \\
Bi-Lvl   &  0.139 & 0.115          & 0.117          & 0.131 & 0.105          & 0.107          & 0.140 & 0.112          & 0.116          \\
MAML     &  0.123 & 0.089          & 0.090          & 0.117 & \textbf{0.073} & 0.076          & 0.130 & 0.083          & 0.085          \\
Meta-Ref &  0.124 & \textbf{0.084} & \textbf{0.084} & 0.117 & \textbf{0.073} & \textbf{0.075} & 0.131 & \textbf{0.080} & \textbf{0.083} \\ \hline
  \end{tabular}
\caption{Average metrics for traffic accident risk estimation on another 90 spatial tasks from test locations.}
\label{tb:perf_regr_2}
\end{table*}

\section{More Visualizations for Comparisons of Candidate Methods}

\begin{figure*}[t]%[h]
    \centering
    \vspace*{-2mm} 
    \hspace*{-6mm} 
    \includegraphics[width=0.95\linewidth]{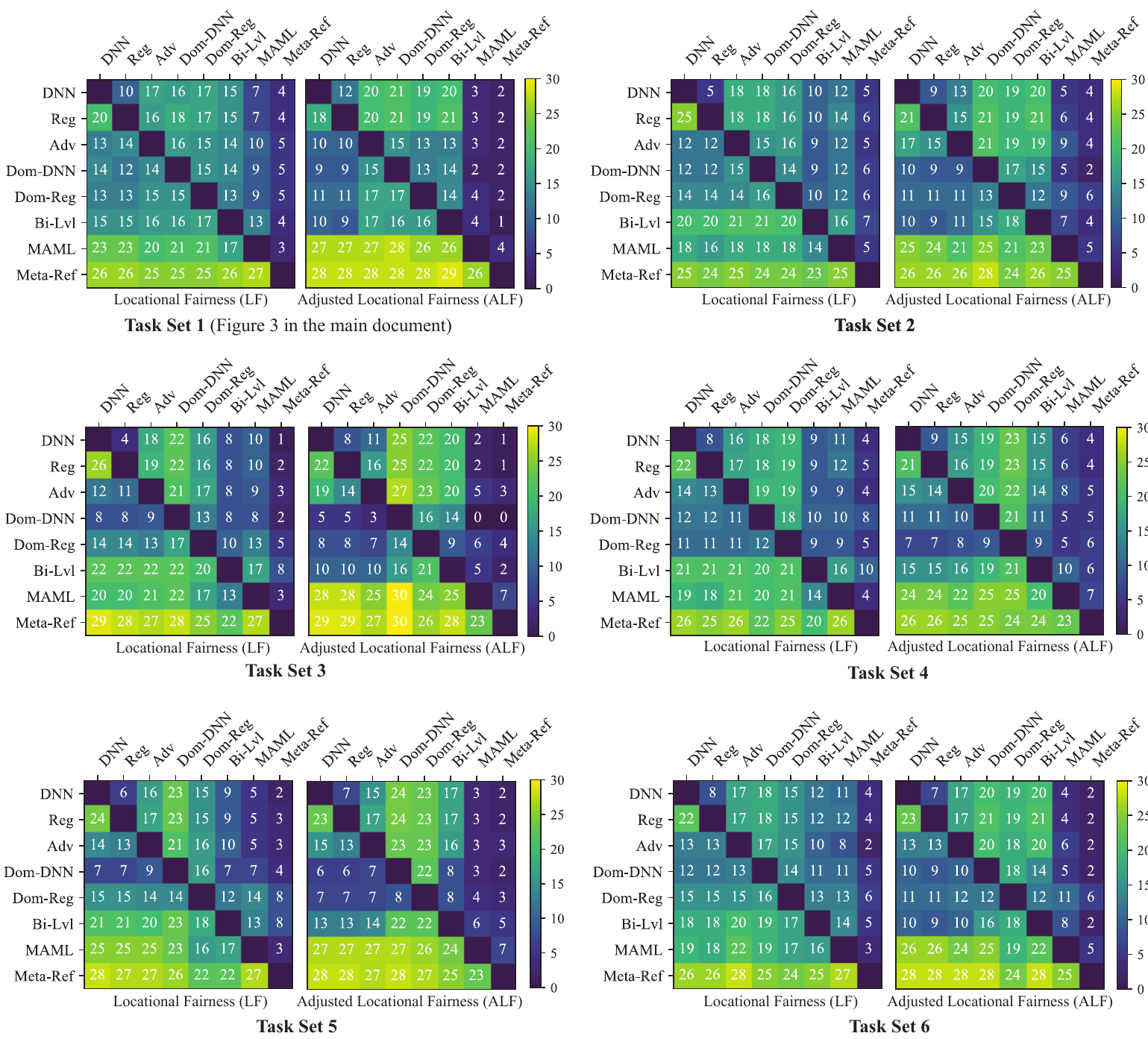}
    \vspace*{-1mm}  
    \caption{
        Pairwise comparison matrices for all methods for crop classification tasks.
    }
    \vspace*{-4mm}
    \label{fig:comparison_mat_more}
\end{figure*}

In the Fig. 3 from the main document, we demonstrated the comparison matrix for the first 30 spatial tasks for the crop classification problem. The rest 150 spatial tasks tested, which we divide into five more spatial tasks (featured in Table 1 of the main document), have exhibited a similar pattern (Fig. \ref{fig:comparison_mat_more} of this document). 
Meta-Ref can deliver better fairness metrics than all baselines in most spatial tasks while maintaining a similar level of prediction performance.

\begin{figure*}[t]%[h]
    \centering
    \vspace*{-2mm} 
    \hspace*{-6mm} 
    \includegraphics[width=0.95\linewidth]{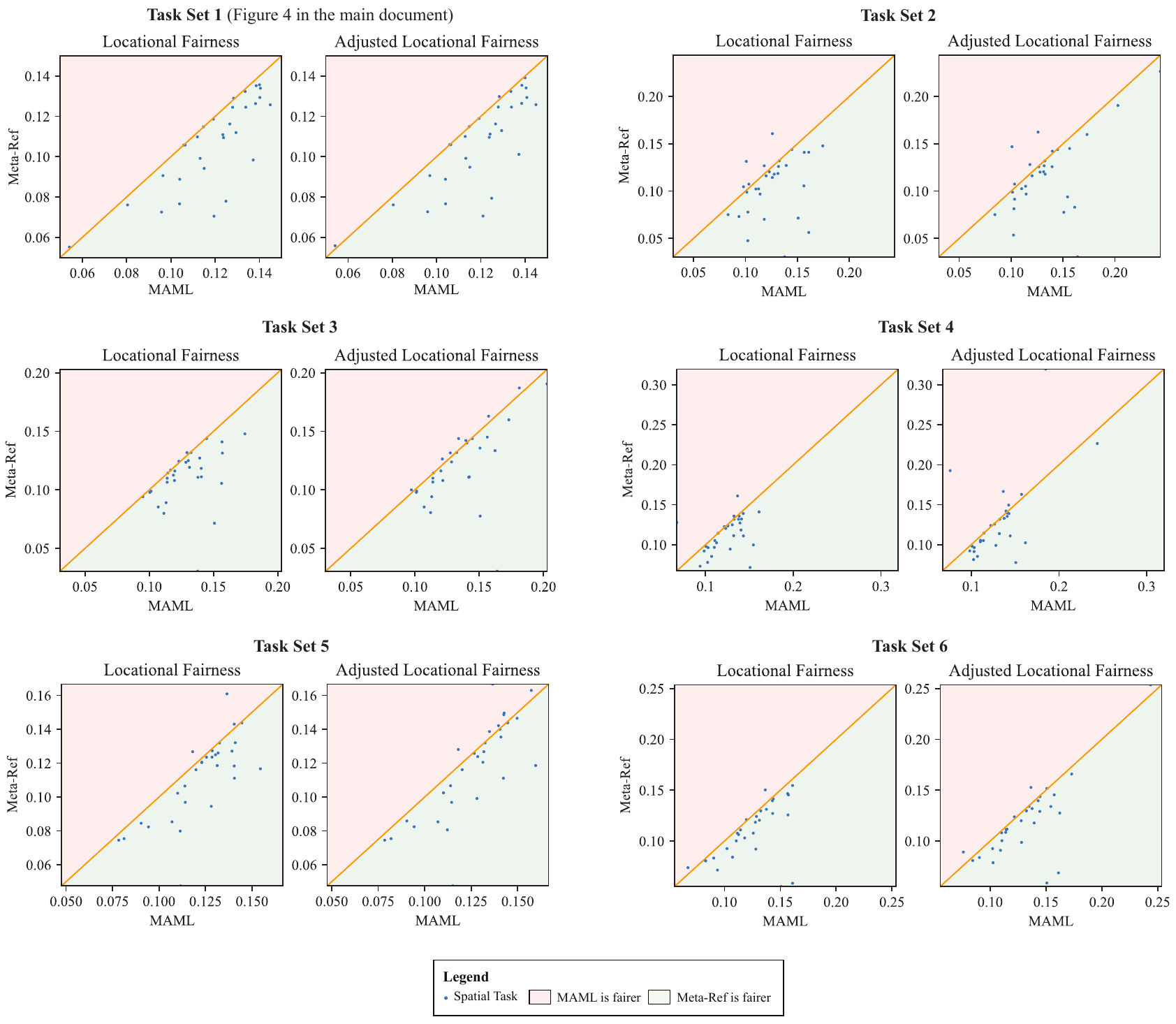}
    \vspace*{-1mm}  
    \caption{
        Comparison between MAML and Meta-Ref on fairness metrics among different spatial tasks for crop classification task sets. For both locational fairness (LF) and adjusted location fairness (ALF), lower values indicate better fairness. Each blue dot represents a spatial task, and the extent of subplots for each task set varies to reflect the range of metrics. 
    }
    \vspace*{-4mm}
    \label{fig:maml_mr_scatter_more}
\end{figure*}

We also demonstrated focused comparisons between MAML and Meta-Ref on one task set in Fig. 4 from the main document. Meta-Ref has shown a competitive edge on the majority of the spatial tasks in each task set. In Fig. \ref{fig:maml_mr_scatter_more}, we provide details on the distribution of LF and ALF metrics across different spatial tasks in the other 5 task sets for this problem that are not displayed in the main document. 

\section{Ablation and Sensitivity Analysis}

Our proposed method includes a three-phase training for Meta-Ref. In the outer loop of the training (in Phase 3) we perform dual meta-update, involving three updates in Eqs. (14, 15, 16). The three updates in Phase 3 of are controlled by three step size hyper-parameters, $\alpha_1$, $\alpha_2$, and $\alpha_3$, respectively. We set the initial value of baseline $\alpha_1 = 10^{-3}$ with expoential decay. In the sensitivity analysis featured in Table \ref{tb:sen_ana} of this document, we link the values of $\alpha_2$ and $\alpha_3$ with $\alpha_1$ by setting a fairness loss scaling factor $\lambda$ so that $\alpha_2 = \alpha_3 = \lambda\cdot\alpha_1$. 
We find that Meta-Ref is robust to this parameter $\lambda$ within the range of [0.01, 10], consistently achieving comparable performance metrics and better fairness scores in both standard and adjusted locational fairness measures than MAML ($\lambda=0$).

To demonstrate the effectiveness of three outer gradient updates in Meta-Ref, we further conduct an ablation study with the following models: (1) \textbf{MR-P2P}: Meta-Ref without applying performance gradient to the prediction model (P2P, Eq. (\ref{eq:perf_pred})); (2) \textbf{MR-F2M}: Meta-Ref without applying fairness gradient to the meta-referee (F2M, Eq. (\ref{eq:fair_mr})); and (3) \textbf{MR-F2P}: Meta-Ref without applying fairness gradient to the prediction model (F2P, Eq. (\ref{eq:fair_pred})). The results of the ablation experiments on crop classification are outlined in Table \ref{tb:abl_ana} of this document. The experiment results suggest that all gradient updates are essential in the training of Meta-Ref. Without P2P, the model fails to generalize with spatial tasks. Without F2M, meta-referee is no longer updated during training, producing arbitrary locational fairness factors that disrupt the training of prediction model. Trimming F2P from Meta-Ref has the smallest impact among three ablation models, since while MR-F2P still underperforms full MAML and Meta-Ref, it has better prediction and location fairness metrics than other baselines.

\begin{table}
\centering
\caption{Sensitivity of Meta-Ref to the fairness loss scaling factor $\lambda$.}
\label{tb:sen_ana}
\begin{tabular}{|c||c|cc|}
\hline
$\lambda$ & F1  & LF & ALF   \\ \hline
0 & 0.718  & 0.122           & 0.122            \\
0.01 & 0.708           & 0.115           & 0.116            \\
0.1  & 0.712           & \textbf{0.112}  & \textbf{0.113}   \\
1    & 0.707           & 0.115           & 0.116            \\
10   & 0.712           & 0.114           & 0.115            \\ \hline         
\end{tabular}
\end{table}

\begin{table}[h]
\centering
\caption{Ablation study results.}
\label{tb:abl_ana}
\begin{tabular}{|c||c|cc|}
\hline
Models &  F1 & LF & ALF   \\ \hline
MAML    & 0.718  & 0.122     & 0.122    \\
MR-P2P  & 0.606          & 0.176     & 0.207     \\
MR-F2M  & 0.645           & 0.143     & 0.161    \\
MR-F2P  & 0.698           & 0.131     & 0.137    \\ 
Meta-Ref  & 0.712  & \textbf{0.112}      & \textbf{0.114}           \\ \hline  
\end{tabular}%
\end{table}

\end{document}